\documentclass{article}

\usepackage[dblblindworkshop, final]{neurips_2025}
\workshoptitle{Embodied World Models for Decision Making}

\usepackage[utf8]{inputenc} 
\usepackage[T1]{fontenc}    
\usepackage{hyperref}       
\usepackage{url}            
\usepackage{booktabs}       
\usepackage{amsfonts}       
\usepackage{nicefrac}       
\usepackage{microtype}      
\usepackage{xcolor}         
\usepackage{graphicx}
\usepackage[labelfont=bf]{caption}

\usepackage{enumitem}
\usepackage{amsmath}
\usepackage{amssymb}
\usepackage{bm}
\usepackage{booktabs}
\usepackage{longtable} 

\usepackage{xcolor}
\usepackage{pifont} 

\newcommand*\circled[1]{\textcircled{\raisebox{-0.9pt}{#1}}}

\title{Learning to Focus: Prioritizing Informative Histories with Structured Attention Mechanisms in Partially Observable Reinforcement Learning}

\author{%
  Daniel De Dios Allegue \space \space \space \space Jinke He \space \space \space \space Frans A. Oliehoek \\
  Delft University of Technology \\
  \texttt{\{ddediosallegue, j.he-4, f.a.oliehoek\}@tudelft.nl}
}

\begin{document}
\maketitle

\begin{abstract}
Transformers have shown strong ability to model long-term dependencies and are increasingly adopted as world models in model-based reinforcement learning (RL) under partial observability. However, unlike natural language corpora, RL trajectories are sparse and reward-driven, making standard self-attention inefficient because it distributes weight uniformly across all past tokens rather than emphasizing the few transitions critical for control. To address this, we introduce structured inductive priors into the self-attention mechanism of the dynamics head: (i) per-head \textbf{memory-length priors} that constrain attention to task-specific windows, and (ii) \textbf{distributional priors} that learn smooth Gaussian weightings over past state–action pairs. We integrate these mechanisms into UniZero, a model-based RL agent with a Transformer-based world model that supports planning under partial observability. Experiments on the Atari 100k benchmark show that most efficiency gains arise from the Gaussian prior, which smoothly allocates attention to informative transitions, while memory-length priors often truncate useful signals with overly restrictive cut-offs. In particular, Gaussian Attention achieves a 77\% relative improvement in mean human-normalized scores over UniZero. These findings suggest that in partially observable RL domains with non-stationary temporal dependencies, discrete memory windows are difficult to learn reliably, whereas smooth distributional priors flexibly adapt across horizons and yield more robust data efficiency. Overall, our results demonstrate that encoding structured temporal priors directly into self-attention improves the prioritization of informative histories for dynamics modeling under partial observability.\footnote{Our code is publicly available in  \href{https://github.com/daniallegue/learning-to-focus}{github.com/daniallegue/learning-to-focus}}

\end{abstract}

\section{Introduction}
Reinforcement learning (RL) \cite{suttonrl} provides a principled framework for sequential decision making, but real-world tasks often violate the Markov assumption and exhibit only partial observability. Such settings are naturally modeled as Partially Observable Markov Decision Processes (POMDPs), which require agents to leverage observation–action histories to reduce uncertainty and achieve robust control \cite{pomdps, planninginpo}.

Model-based RL addresses this challenge by learning an explicit world model of environment dynamics \cite{suttonrl}, which can be used to plan or imagine future trajectories. A seminal example is MuZero \cite{muzero}, which learns a joint representation, dynamics, and value model in latent space, paired with Monte Carlo Tree Search \cite{mcts} to achieve state-of-the-art performance in board games and Atari. More recently, UniZero \cite{unizero} replaced MuZero’s recurrent dynamics with a Transformer backbone, using masked self-attention to capture long-range dependencies in latent state–action sequences and improve sample efficiency under partial observability.

Despite this architectural shift, UniZero often remains sample-inefficient in low-data regimes because it inherits assumptions from natural language modeling: namely, that sequential data are abundant, balanced, and richly interdependent. In reality, RL trajectories consist of long stretches of uninformative transitions, sparse rewards, skewed return distributions, and a limited number of interactions \cite{rltrajectories, her}. In Transformer-based world models, standard self-attention treats all past tokens within the history as equally relevant, making it hard to identify the sparse transitions that actually drive reward. Unlike language modeling, where vast corpora make even rare dependencies learnable \cite{lmfewshot}, RL agents operate on scarce and noisy trajectories, requiring attention mechanisms that explicitly prioritize informative segments of history \cite{transformersinrl}.

To address this limitation, we enhance the UniZero world model by introducing two structured temporal priors into the self-attention layers of its dynamics head. The dynamics head predicts the next latent state $z_{t+1}$ and immediate reward $r_t$ based on attention-weighted histories. The first prior, a \textbf{memory-length prior}, restricts each attention head to a learnable contiguous window, approximating the minimal history required for accurate prediction. The second, a \textbf{distributional prior}, applies smooth Gaussian weighting over past tokens, emphasizing those most informative for immediate outcomes. We instantiate these as \textbf{Adaptive Attention} (memory-length prior), \textbf{Gaussian Attention} (distributional prior), and their combination, \textbf{Gaussian Adaptive Attention}.

On the Atari 100k benchmark, Gaussian Attention yields a 77\% relative improvement in human-normalized mean score over UniZero’s standard self-attention. This gain stems from its ability to allocate weight smoothly across past transitions, capturing relevant temporal dependencies without imposing sharp cutoffs. In contrast, Adaptive Attention often misestimates the true dependency horizon, either truncating important signals or including irrelevant ones, reducing sample efficiency. Combining the two mechanisms degrades performance: the hard span mask truncates Gaussian tails, negating its smooth weighting benefits. These results highlight a general guideline for model-based RL under partial observability: smooth distributional priors offer more robust and data-efficient dynamics modeling than rigid memory-length priors.

Our contributions are as follows:
\begin{itemize}
\item We propose two structured temporal priors for self-attention in world models: a memory-length prior enforcing per-head learnable look-back windows, and a distributional prior introducing smooth Gaussian weightings over histories.
\item We integrate these mechanisms into the UniZero agent and demonstrate on Atari 100k that Gaussian Attention achieves substantial gains in human-normalized mean and median scores, with negligible computational overhead.
\item We analyze the complementary behavior of hard and smooth priors, showing how Gaussian priors reliably capture diverse temporal dependencies while memory-length priors offer benefits in limited cases.
\item Through systematic ablations across Atari games, we isolate the effects of each prior and its regularization, confirming robustness to initialization and low additional computational cost.
\end{itemize}

\section{Background}
\textbf{MDPs and POMDPs.}  
A Markov Decision Process (MDP) is defined by the tuple $(\mathcal{S}, \mathcal{A}, P, R, \gamma)$, where $\mathcal{S}$ is the state space, $\mathcal{A}$ is the action space, $P(s'\mid s,a)$ denotes the transition probability, $R(s,a)$ is the reward function, and $\gamma \in [0,1)$ is the discount factor \cite{suttonrl}. The agent seeks a policy $\pi : \mathcal{S} \to \mathcal{A}$ that maximizes the expected discounted return $E[\sum_{t=1}^{\infty} \gamma^t R(s_t, a_t)]$, satisfying the Bellman optimality equation. A Partially Observable MDP (POMDP) extends this formulation with an observation space $\mathcal{O}$ and observation probabilities $O(o \mid s,a)$, since the true state is not directly observable, and thus is defined by $(\mathcal{S}, \mathcal{A}, \mathcal{O}, P, R, O, \gamma)$. To act under partial observability, the agent maintains a belief distribution $b$ over states, updated after action $a$ and observation $o$ as $b_{a,o}(s') \propto O(o \mid s',a)\sum_s P(s'\mid s,a)b(s)$. Not all observations are equally informative, and a central objective in planning under partial observability is to identify a minimal subset of history sufficient for predicting future transitions and rewards. Influence-Based Abstraction (IBA) formalizes this by identifying d-separating observation sets that render the future conditionally independent of the remaining history \cite{iba}, echoing state-abstraction principles in RL \cite{equivalencemdp}.  


\textbf{Transformers.}  
Transformers \cite{attentionisall} have emerged as powerful alternatives to recurrent neural networks (RNNs) for long-sequence modeling. Given an input sequence of length $N$, each token is projected into queries $Q \in \mathbb{R}^{N \times d_k}$, keys $K \in \mathbb{R}^{N \times d_k}$, and values $V \in \mathbb{R}^{N \times d_v}$. Self-attention then aggregates contextual information via:  
\begin{equation}\label{attention}
    \mathrm{Attention}(Q,K,V) = \mathrm{softmax}\!\Bigl(\tfrac{QK^\top}{\sqrt{d_k}}\Bigr)V.
\end{equation}  
Since self-attention is permutation-invariant, positional encodings, either fixed or learned, are added to token embeddings to inject order information \cite{bert}.
By combining global context aggregation with positional encodings, Transformers effectively capture long-range dependencies that truncated RNNs fail to model \cite{transformerxl}. This has made Transformer architectures compelling candidates for world models in RL, where long-horizon planning and memory are critical \cite{transformersinrl, transformerworldmodel}.  

\textbf{MuZero.}  
MuZero \cite{muzero} achieves superhuman performance in board games and Atari by integrating Monte Carlo Tree Search (MCTS) with a learned latent dynamics model. At each time step $t$, MuZero employs:  
\begin{enumerate}
  \item \textbf{Encoder:} $z^0_t = h_\theta(o_{1:t})$, mapping the observation history to a latent state.  
  \item \textbf{Dynamics head:} $(z_{t+1}, r_t) = g_\theta(z_t, a_t)$, unrolling latent states and predicting rewards recurrently.  
  \item \textbf{Prediction head:} $(\pi_t, v_t) = f_\theta(z_t)$, producing policy logits and value estimates.  
\end{enumerate}  
Although powerful, MuZero’s recurrent dynamics suffer from vanishing gradients and a fixed unroll horizon, which limit its ability to capture long-range dependencies \cite{vanishinggradients}.  

\begin{figure}[!t]
    \centering
    \includegraphics[width=1\linewidth]{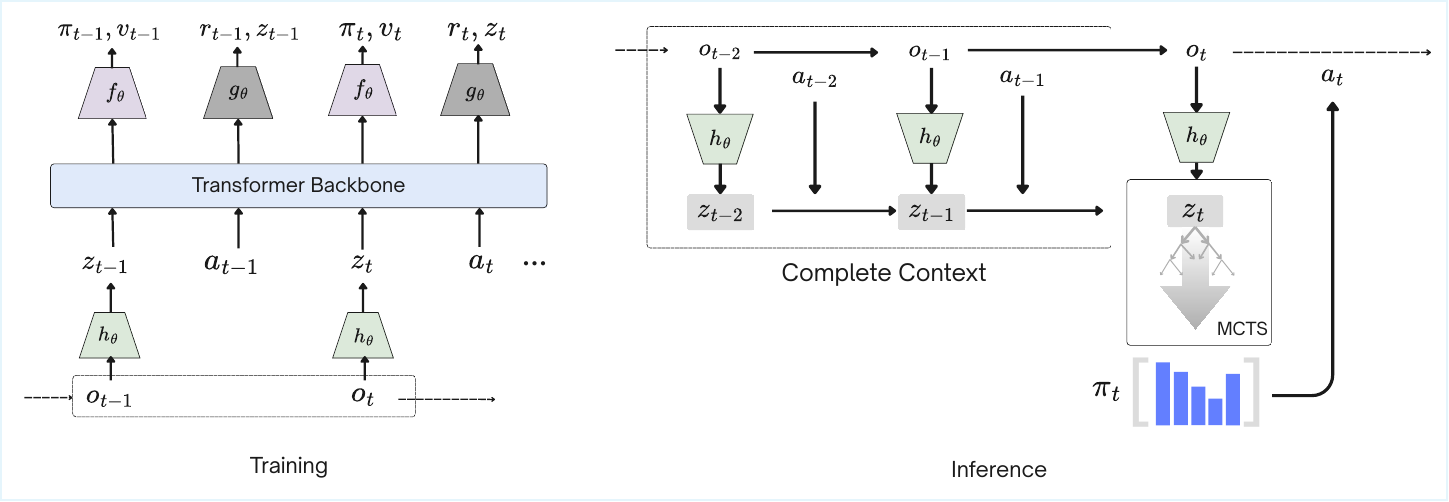}
\caption{\textbf{UniZero architecture.} Overview of the UniZero world model $W = (h_\theta, g_\theta, f_\theta)$. The encoder $h_\theta$ maps observations $o_t$ to latent states $z_t$. The dynamics head $g_\theta$ (a Transformer with masked self-attention) consumes past latent state-action pairs to predict the next state $z_{t+1}$ and reward $r_t$. The prediction head $f_\theta$ outputs the policy $\pi_t$ and value $v_t$ for MCTS planning.}
    \label{fig:architecture}
\end{figure}

\textbf{UniZero.}  
UniZero \cite{unizero} retains MuZero’s overall world model tuple $W = (h_\theta, g_\theta, f_\theta)$ but parameterizes $g_\theta$ and $f_\theta$ with a Transformer backbone. Unlike MuZero, whose encoder produces only a single latent state summarizing the entire history, UniZero encodes each observation individually as $z_i = h_\theta(o_i)$, yielding a sequence $z_{1:t}$. The sequence of observation–action pairs $\bigl[(z_1,a_1), \dots, (z_t,a_t)\bigr]$ is then processed by $L$ stacked Transformer layers, each with $h$ attention heads. Masked self-attention ensures that token $i$ attends only to past tokens, preventing future leakage (see \autoref{fig:architecture}).

The outputs from all heads are concatenated and projected through a final linear layer, integrating the diverse subspaces captured by each head. This allows UniZero to capture dependencies far beyond MuZero’s fixed horizon, though at quadratic complexity in the sequence length, the number of layers $L$, and the number of heads $h$. Moreover, because self-attention initially treats all past tokens as equally relevant, the model must learn relevance weights during training, often leading to sample inefficiency.  

The final Transformer layer outputs the next latent state $z_{t+1}$ and immediate reward $r_t$, which are passed to the unchanged prediction head $f_\theta$ to produce $\pi_t$ and $v_t$. UniZero, like MuZero, is trained via joint model–policy optimization, maintaining a soft-target world model $\hat{W} = (\hat h_\theta, \hat g_\theta, \hat f_\theta)$ to stabilize learning \cite{joint-training}. By leveraging global temporal context, UniZero improves long-horizon performance, but its uniform attention weighting motivates the structured temporal priors we introduce in this work.

\section{Related Work}

\noindent\textbf{RL in POMDPs.}  
Under partial observability, model-free methods typically rely on recurrent networks to infer hidden states \cite{deepqlearningpomdp}, whereas model-based approaches learn latent world models for planning. Early frameworks such as predictive state representations \cite{predictive} have evolved into deep generative models such as Dreamer, which combine variational inference and recurrent state-space models to compactly represent belief states and enable efficient long-horizon planning \cite{worldmodels,dreamer}.  

\noindent\textbf{Memory Mechanisms in DRL.}  
Many deep RL methods explicitly incorporate memory to handle partial observability. Simple approaches stack the last $k$ frames \cite{dqn}, recurrent architectures summarize the entire action–observation history into fixed-size states \cite{deepqlearningpomdp}, and external differentiable memories further expand capacity but often introduce training instability \cite{diffmemory}. Influence-Aware Memory (IAM), inspired by Influence-Based Abstraction (IBA) \cite{iba}, learns gating mechanisms that selectively retain past observations predictive of future outcomes \cite{iba-memory}.  

\noindent\textbf{Transformer-based World Models.}  
Recent Transformer adaptations in RL leverage self-attention to capture long-range dependencies, but most do not incorporate inductive priors tailored to RL sequences. On the model-free side, methods such as GTrXL and Transformer-XL stabilize attention via gating and relative encodings \cite{stabilising,transformerxl}, Decision Transformer reframes control as return-conditioned masked attention over past trajectories \cite{decisiontransformer}, and Adaptive Span Transformer reduces computation by learning per-head context lengths without building an explicit dynamics model \cite{adaptiveinrl}. On the model-based side, hybrids such as IRIS \cite{iris} and TransDreamer \cite{transdreamer} integrate Transformers into latent world models, rolling out imagined trajectories for planning to achieve strong sample efficiency.  However, most existing Transformer-based world models in RL rely on fixed or NLP-inspired positional encodings (e.g., sinusoidal or relative embeddings), which emphasize computational efficiency rather than task relevance. In contrast, we introduce structured temporal priors to better align attention with reward-relevant dependencies.

\section{Dynamics Modelling with Self-Attention Priors}

\begin{figure*}[!t]
  \centering
  \includegraphics[width=\textwidth]{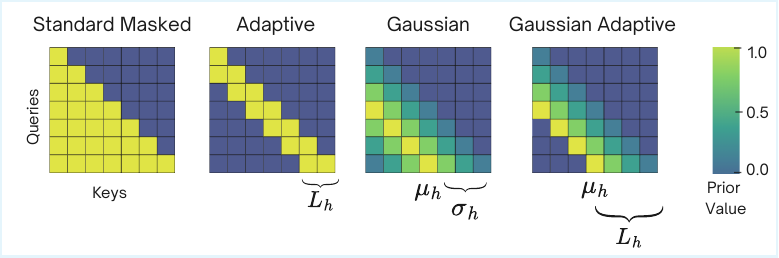}
    \caption{\textbf{Attention priors.} Visualization of the attention prior matrices described in Eqs.~(5)--(9). Rows correspond to queries $i$ (current timestep) and columns to keys $j$ (past tokens). (From left to right) standard causal masking, the memory-length prior (Eq.~\ref{eq:adaptive_mask}), the Gaussian distributional prior (Eq.~\ref{eq:gaussian_mask}), and their combination (Eq.~\ref{eq:gaussian_adaptive}). Yellow indicates high prior bias, dark blue zero bias. Together, these priors constrain $\alpha_{ij}$ to emphasize reward-relevant temporal context $c_t^{(h)}$.}
  \label{fig:method}
\end{figure*}

In UniZero’s world model, the dynamics function aggregates past latents and actions up to time $t$ into a history $h_t$, and predicts the next latent and reward:
\begin{equation}
    (\hat{z}_{t+1},\hat{r}_t) = g_\theta(z_{\le t},a_{\le t}) = g_\theta(h_t),
\end{equation}
where relevance is computed via self-attention with weights $\{\alpha_{ij}\}_{j=1}^i$ (with $i$ the current query and $j$ the key). Under partial observability, however, only a limited window of context and a sparse set of key events truly drive accurate predictions. To better align attention with these reward-relevant dependencies, we introduce two structured temporal priors into the attention mechanism: (i) a \textbf{memory-length prior} that enforces a learnable finite look-back span, and (ii) a \textbf{distributional prior} that softly emphasizes tokens according to a Gaussian saliency distribution.  
Our goal is to bias self-attention toward histories that matter most for predicting dynamics and rewards, thereby improving sample efficiency in low-data, partially observable RL settings. \autoref{fig:method} illustrates the different attention priors described in this section.

\subsection{Memory-Length Prior}
Many partially observable environments admit a finite effective memory: only the most recent $n$ steps are needed to predict the next reward \cite{predictive,a3c}. Imposing this prior focuses the model on a minimal history window, reducing redundant computation over distant tokens. Formally,
\begin{equation}\label{eq:memory_len}
  \mathbb{E}[r_t \mid h_{1:t}, a_t]
  \;=\;
  \mathbb{E}[r_t \mid h_{t-n+1:t}, a_t].
\end{equation}

We implement this using \textbf{Adaptive Attention} \cite{adaptiveattentionspan}. Each head $h$ learns a scalar parameter $s_h$, transformed via softplus into a positive span $L_h = \mathrm{softplus}(s_h)$. A hard mask over relative positions is then constructed:
\begin{equation}\label{eq:adaptive_mask}
  M^{(h)}_{ij} =
  \begin{cases}
    0, & i - j \le L_h,\\
    -\infty, & i - j > L_h,
  \end{cases}
\end{equation}
so that queries at $i$ can only attend within their learned look-back span. Attention weights become
\begin{equation}
  \mathrm{Attention}^{(h)}
  = \mathrm{softmax}\!\bigl(\tfrac{Q^{(h)}K^{(h)\top}}{\sqrt{d_k}} + M^{(h)}\bigr).
\end{equation}


To prevent trivial solutions where all spans grow without bound, we apply an $\ell_1$ penalty, encouraging the model to learn minimal but sufficient spans \cite{lasso,equivalencemdp}. Each attention head $h$ produces a context vector
\begin{equation}
c_t^{(h)} = \sum_{j=1}^t \alpha_{tj}^{(h)} [z_j; a_j],
\end{equation}
where $\alpha_{tj}$ are the learned attention weights that determine the relative importance of past steps when computing the context, and $[z_j; a_j]$ denotes the concatenated latent state and action at step $j$. By constraining spans $L_h$, different heads specialize at distinct temporal scales, yielding a multi-scale representation when their context vectors are combined into $h_t$.

\subsection{Distributional Prior}

In partially observable settings, only a sparse subset of tokens carries predictive signal for $(\hat z_{t+1}, \hat r_t)$. We capture this \emph{distributional prior} by learning a Gaussian positional kernel.

Each head $h$ learns parameters $\mu_h, \sigma_h > 0$, defining
\begin{equation}\label{eq:gaussian_mask}
  G^{(h)}_{ij} = -\frac{(i - j - \mu_h)^2}{2\sigma_h^2}.
\end{equation}
This is added to the scaled dot-product logits:
\begin{equation}
  \mathrm{Attention}^{(h)}
  = \mathrm{softmax}\!\Bigl(\tfrac{Q^{(h)}K^{(h)\top}}{\sqrt{d_k}} + G^{(h)}\Bigr),
\end{equation}
so that queries at $i$ privilege tokens at offset $\mu_h$ with sharpness $\sigma_h$ \cite{gaam}. Unlike spans, $\mu_h$ and $\sigma_h$ are unconstrained: $\sigma_h$ may expand to broad attention or shrink to narrow focus, thereby giving each head a smooth, learned saliency profile through $G^{(h)}$. Different heads capture different offsets and spreads, producing complementary temporal filters that are concatenated into $h_t$.

\subsection{Combining Priors}
Finally, we combine the two priors by defining
\begin{equation}\label{eq:gaussian_adaptive}
  B^{(h)}_{ij} = G^{(h)}_{ij} + M^{(h)}_{ij},
\end{equation}
and apply it within the attention mechanism by adding $B^{(h)}$ as a bias term to the scaled dot-product before the softmax. This \textbf{Gaussian Adaptive Attention} enforces a finite horizon while retaining smooth saliency within it, thereby combining the strengths of memory-length and distributional priors.



\section{Experiments}
We evaluate our Transformer-based world model augmented with attention priors on the Atari 100k benchmark, a widely used testbed for sample efficiency in model-based reinforcement learning. This suite spans diverse reward densities, horizon lengths, and stochastic dynamics. Our evaluation considers both aggregate performance and the contribution of each prior through controlled ablations.  

\subsection{Experimental Setup}
Agents are trained on 26 Atari environments for $100k$ steps, with performance averaged over five random seeds ($1$–$5$) and reported as human-normalized scores following \cite{atari100k}. We utilize Single-Task (ST) training (separate model per environment) to isolate the effects of attention priors. Unless noted, we adopt UniZero's default hyperparameters from \citet{unizero}. Full details, configurations, and reproduction instructions are in \autoref{appendix:reproducibility}.

\begin{table*}[!ht]
  \centering
  \captionsetup{skip=8pt}
  \scriptsize
  \setlength{\tabcolsep}{4pt}    
  \renewcommand{\arraystretch}{1.2}
  \caption{\textbf{Raw Atari 100k scores} comparing our attention‐biased UniZero variants against reproduced UniZero and MuZero baselines. MuZero results are from LightZero reproductions in \cite{unizero} (three seeds), while Random and Human scores are from \cite{unizero}. All “Ours” results are averaged over five seeds. \textbf{Bold} entries denote the superior method between the \textit{UniZero ST} baseline and our attention-biased methods, while \underline{underlined} values indicate the overall best-performing method.}

  \label{tab:atari-results}
  \resizebox{\textwidth}{!}{
      \begin{tabular}{l@{\quad}*{7}{r@{\quad}}}
        \hline
        Game            & Random      & Human       &  MuZero  & UniZero ST (Baseline) & Adaptive UniZero (Ours) & Gaussian UniZero (Ours) & Gaussian Adaptive UniZero (Ours) \\
        \hline
        Alien           & $227.8$     & $7127.7$    &  $300.0$    & $468.5$             & \underline{$\bm{570.6}$}         & $483.3$              & $509.6$                      \\
        Amidar          & $5.8$       & $1719.5$    &  \underline{$90.0$}     & $57.2$            & $57.9$               & $\bm{71.2}$          & $53.4$                       \\
        Assault         & $222.4$     & $742.0$     &  \underline{$609.0$}           & $341.9$             & $423.5$                & $\bm{486.8}$           & $333.7$                        \\
        Asterix         & $210.0$     & $8503.7$    & \underline{$1400.0$}                        & $495.3$             & $500.1$                & $\bm{619.9}$           & $333.6$                        \\
        BankHeist       & $14.2$      & $753.1$     & \underline{$223.0$}                & $91.3$              & $13.3$                 & $\bm{165.1}$           & $0.7$                        \\
        BattleZone      & $2360.0$    & $37187.5$   &  \underline{$7587.0$} & $\bm{6000.0}$       & $5872.5$               & $5361.6$               & $5297.6$                       \\
        Boxing          & $0.1$       & $12.1$      & \underline{$20.0$}          & $0.1$               & $-9.5$                 & $\bm{2.4}$             & $-11.3$                        \\
        Breakout        & $1.7$       & $30.5$      & $3.0$     & $3.7$                   & $0.8$                    & \underline{$\bm{5.1}$}                      & $0.5$                              \\
        ChopperCommand  & $811.0$     & $7387.8$    & $1050.0$ & $1169.0$            & $872.5$                & \underline{$\bm{1263.4}$}          & $735.2$                        \\
        CrazyClimber    & $10780.5$   & $35829.4$   &  \underline{$22060.0$}         & $7418.9$            & $4326.6$               & $\bm{7966.6}$          & $2020.0$                       \\
        DemonAttack     & $152.1$     & $1971.0$    &  \underline{$4601$}                 & $236.3$             & $187.4$                & $\bm{267.0}$                & $166.4$                        \\
        Freeway         & $0.0$       & $29.6$      &  \underline{$12.0$}    & $0.0$                  & $0.7$                  & $0.1$                  & $\bm{2.6}$                      \\
        Frostbite       & $65.2$      & $4334.7$    & $260.0$      & $239.8$             & \underline{$\bm{261.2}$}                & $236.7$                & $162.2$                        \\
        Gopher          & $257.6$     & $2412.5$    & $346.0$          & $606.7$             & $646.4$                & \underline{$\bm{798.8}$}           & $240.0$                         \\
        Hero            & $1027.0$    & $30826.4$   & \underline{$3315.0$} & $1483.0$            & $1422.2$               & $699.6$                & $\bm{2414.4}$                  \\
        Jamesbond       & $29.0$      & $302.8$     & $90.0$   & $201.7$             & $156.7$                & \underline{$\bm{362.0}$}           & $75.9$                         \\
        Kangaroo        & $52.0$      & $3035.0$    & $200.0$       & $842.6$             & $488.6$                & \underline{$\bm{1636.4}$}          & $367.9$                        \\
        Krull           & $1598.0$    & $2665.5$    &  \underline{$5191.0$}             & $2539.8$            & $2647.5$               & $\bm{3108.8}$          & $1964.0$                       \\
        KungFuMaster    & $258.5$     & $22736.3$   &  $6100.0$             & $2019.0$            & $8546.5$               & \underline{$\bm{9424.5}$}          & $644.3$                        \\
        MsPacman        & $307.3$     & $6951.6$    &  $1010.0$                     & $643.9$             & \underline{$\bm{1103.3}$}          & $726.6$                & $394.7$                        \\
        Pong            & $-20.7$     & $14.6$      & $-15.0$                     & $-14.5$             & $-19.6$                 & \underline{$\bm{-7.1}$}            & $-20.3$                       \\
        PrivateEye      & $24.9$      & $69571.3$   & \underline{$100.0$} & $\bm{93.3}$         & $-60.1$                & $57.6$                 & $80.0$                         \\
        Qbert           & $163.9$     & $13455.0$   &  $1700.0$         & $677.2$             & $941.5$                & \underline{$\bm{1741.8}$}          & $356.3$                        \\
        RoadRunner      & $11.5$      & $7845.0$    &  \underline{$4400.0$} & $1941.3$            & $\bm{2164.5}$          & $1948.4$               & $1400.0$                        \\
        Seaquest        & $68.4$      & $42054.7$   & $466.0$ & $384.1$             & $293.2$                & \underline{$\bm{485.7}$}           & $ 273.3$                        \\
        UpNDown         & $533.4$     & $11693.2$   & $1213.0$   & $2018.0$            & $1374.7$               & \underline{$\bm{2373.8}$}          & $1246.4$                       \\
        \hline
        Normalized Mean   & $0.000$ & $1.000$ & \underline{$0.44$} & $0.13$ & $0.095$ & $\bm{0.23}$ & $0.00$ \\
        Normalized Median & $0.000$ & $1.000$ & \underline{$0.13$} & $0.05$ & $0.05$ & $\bm{0.10}$ & $0.02$ \\
        \hline
      \end{tabular}
      }
\end{table*}

\noindent \textbf{Attention Prior Initialization.}  
We initialize all attention priors to align with typical temporal dependencies in Atari trajectories. For Adaptive Attention, each head begins with a span of $L_h^0 = 0.3\,L_{\max} \approx 6$, following the recommendations of \citet{adaptiveinrl}. Gaussian Attention is initialized with mean offset $\mu_h = 6$ and standard deviation $\sigma_h = 1$, while Gaussian Adaptive Attention learns both $\mu_h$ and $\sigma_h$ but applies a hard cutoff at $L_h^0 = 10$. To ensure comparable starting conditions, initial distributional logits are sampled from $\mathcal{N}(\mu_h, \sigma_h^2)$, exactly matching the Gaussian prior.  

\noindent \textbf{Baselines.}  
We compare against two established model-based RL baselines implemented in the LightZero framework \cite{lightzero}: (i) \emph{MuZero} \cite{muzero}, which combines latent dynamics with Monte Carlo Tree Search, and (ii) \emph{UniZero} \cite{unizero}, which replaces MuZero’s recurrent core with a Transformer backbone. Both baselines are trained for 100k steps per environment under identical hyperparameters, ensuring that performance differences arise solely from the proposed priors.  

\subsection{Performance Results}
\autoref{tab:atari-results} reports Atari 100k results against UniZero (ST) and MuZero. Gaussian UniZero delivers the best overall performance, improving HNS from $0.13$ to $0.23$ (+77\%) and HMS from $0.05$ to $0.10$ (+100\%), outperforming the baseline in 19 of 26 games. Adaptive and Gaussian Adaptive variants yield inconsistent or weaker results, with Adaptive only matching the baseline on HMS. Overall, smooth Gaussian priors provide consistent sample-efficiency gains, while rigid span cutoffs hurt performance. Full learning curves can be found in \autoref{appendix:learned}.
.

Gaussian Attention consistently outperforms alternatives because it distributes weights smoothly across short- and mid-range temporal offsets, effectively capturing both immediate and moderately delayed dependencies \cite{transformersinrl}. By contrast, Adaptive Attention’s hard spans often misestimate the relevant horizon, either truncating delayed yet informative signals or incorporating irrelevant context. Combining Gaussian weighting with a hard cutoff further degrades performance: truncating the Gaussian kernel removes useful tails and produces conflicting priors. Together, these findings suggest a general guideline for model-based RL under partial observability: smooth, learnable positional priors offer a more robust and flexible mechanism for temporal modeling than rigid memory windows. Future directions include extending Gaussian priors to multi-task settings, where shared temporal structure across games could further improve generalization.  

\subsection{Ablation Studies}
To isolate the contributions of each prior, we conduct ablations on four representative Atari games: Pong, MsPacman, Jamesbond, and Freeway, which span diverse observation complexities, reward structures, and temporal dependencies.  

\noindent\textbf{Regularization Ablation.}  
We compare three penalties on the learned span vector $L_h$, each with penalty coefficient $\lambda = 0.025$ as in \citet{adaptiveinrl}:  
\begin{enumerate}[label=\protect\circled{\arabic*}]
    \item Max-norm $\ell_{\text{max}}$: enforces $\|L_h\|_\infty \leq c$, restricting each head to the most recent tokens \cite{maxnorm}.
    \item $\ell_1$: adds $\lambda \sum_j L_{h,j}$, encouraging sparsity by driving many spans to zero while letting a few grow.
    \item $\ell_2$: adds $\lambda \sum_j L_{h,j}^2$, softly shrinking spans while preserving long-range context.
\end{enumerate}  

\begin{figure}[h!]
    \centering
    \includegraphics[width=0.95\linewidth]{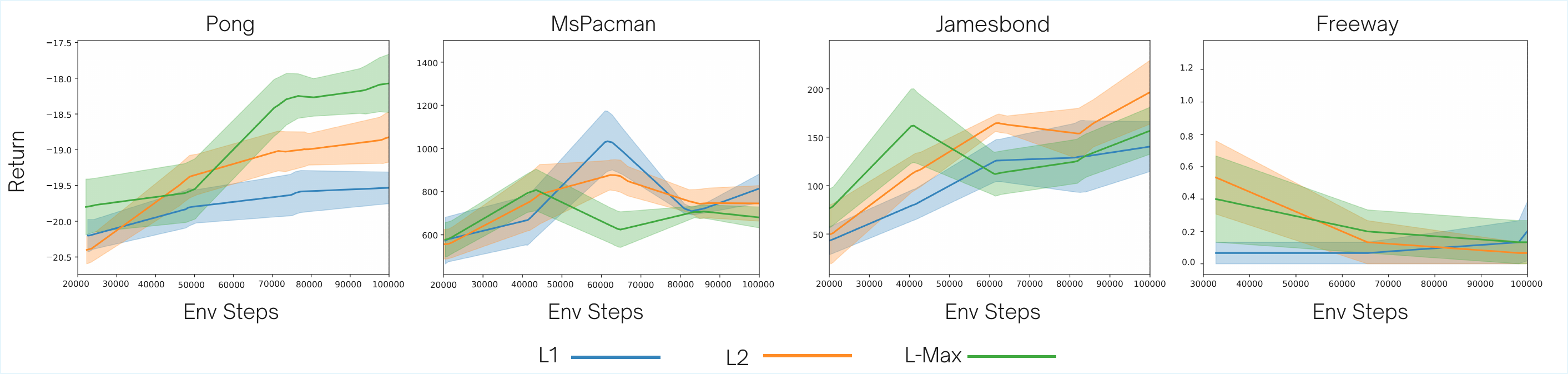}
    \caption{\textbf{Regularization ablation.} Comparison of $\ell_1$ (blue), $\ell_2$ (orange), and max-norm (green) penalties across four Atari games over five seeds. Shaded regions denote standard error. Each scheme exhibits task-specific strengths, but $\ell_2$ achieves the most consistent performance overall.}
    \label{fig:reg-ablation}
\end{figure}

In practice, max-norm favors purely short-term attention; $\ell_1$ produces a bimodal mix of very short and very long spans; and $\ell_2$ encourages balanced recency while retaining moderate long-range dependencies. \autoref{fig:reg-ablation} illustrates these effects: max-norm performs best in short-horizon tasks, $\ell_2$ dominates in mid-horizon settings, and $\ell_1$ occasionally excels in long-horizon environments by retaining sparse but wide spans. Overall, $\ell_2$ generalizes most robustly, striking a balance between stability and flexibility.  

\noindent\textbf{Initial Parameter Sensitivity.}  
As shown in \autoref{tab:init-param-ablation}, sensitivity analysis reveals robustness to initial spans ($L_h$) and Gaussian centers ($\mu_h$), which adapt quickly. However, the Gaussian width $\sigma_h$ is critical: narrower priors ($\sigma_h = 1$) consistently yield superior results, whereas wider priors ($\sigma_h = 3$) underperform. This indicates that a tight positional prior is essential for a strong inductive initial distribution.

\begin{table}[h!]
  \centering
  \scriptsize
  \captionsetup{skip=8pt}
  \setlength{\tabcolsep}{3pt}
  \renewcommand{\arraystretch}{0.85}
    \caption{\textbf{Ablation on initialization of attention priors.} Mean $\pm$ standard error over five seeds on four Atari games. Varying initial spans $L_h$ or offsets $\mu_h$ has little effect, while narrow Gaussian widths ($\sigma_h = 1$) consistently improve performance. \textbf{Bold} entries mark the best result per game.}
  \begin{tabular*}{\columnwidth}{@{\extracolsep{\fill}} l c c c c }
    \toprule
     & \textbf{Pong} & \textbf{MsPacman} & \textbf{Jamesbond} & \textbf{Freeway} \\
    \midrule
    $L_h = 2$   & $-18.5\pm0.4$   & $716.7\pm58.9$       & $180.0\pm49.8$     & $2.2\pm2.1$      \\
    $L_h = 6$   & $-19.6\pm0.4$   & $\bm{1103.3\pm345.8}$ & $156.7\pm29.8$     & $0.7\pm0.6$      \\
    $L_h = 10$  & $-18.7\pm0.5$   & $633.3\pm47.4$       & $130.0\pm34.7$     & $\bm{2.7\pm2.1}$ \\
    \midrule
    $\mu_h = 2$   & $\bm{-6.9\pm1.8}$   & $805.3\pm112.6$      & $293.3\pm49.6$     & $0.0\pm0.0$      \\
    $\mu_h = 6$   & $-7.9\pm1.0$       & $726.7\pm98.2$       & $\bm{362.1\pm53.1}$ & $0.1\pm0.1$      \\
    $\mu_h = 10$  & $-10.5\pm1.0$      & $894.7\pm101.8$      & $290.0\pm58.4$     & $0.1\pm0.1$      \\
    \midrule
    $\sigma_h = 1$   & $-7.9\pm1.0$       & $726.7\pm98.2$       & $\bm{362.1\pm53.1}$ & $0.1\pm0.1$      \\
    $\sigma_h = 3$   & $-15.1\pm0.7$      & $638.7\pm47.6$       & $196.7\pm24.4$     & $0.0\pm0.0$      \\
    \bottomrule
  \end{tabular*}
  \label{tab:init-param-ablation}
\end{table}


\noindent\textbf{Limitations.}  
Our evaluation is restricted to Atari, leaving open whether the proposed attention priors generalize to continuous-control or multi-task settings. In addition, the learned look-back spans require regularization to avoid collapse to trivial extremes, which may limit adaptability in environments with highly variable temporal dependencies. Future work should investigate more flexible temporal priors and evaluate their robustness across broader RL domains, including continuous-control benchmarks such as \cite{dmcontrol}.





\section{Conclusion}

In NLP, Transformers benefit from massive, balanced corpora where long‐range dependencies recur frequently, allowing self‐attention to capture them implicitly. In contrast, model-based RL agents must identify the few reward-relevant dependencies hidden within sparse and correlated trajectories under limited supervision. This mismatch makes standard self-attention sample-inefficient, as it spreads its focus across many uninformative transitions rather than concentrating on the critical ones. We addressed this by incorporating two inductive priors into UniZero’s dynamics head: a \textbf{memory‐length prior}, restricting each head to a finite span, and a \textbf{distributional prior}, implemented as a smooth Gaussian positional prior.  

Experiments on Atari‐100k demonstrate that Gaussian positional priors substantially improve sample efficiency, delivering a 100\% relative gain in human‐normalized median score, while hard span cutoffs degrade performance by truncating delayed yet informative signals. These results suggest a broader principle: smooth, learnable temporal priors align better with the irregular dependency structure of RL trajectories than rigid memory windows. Looking ahead, structured temporal priors in self-attention promise to improve robustness and data efficiency in Transformer world models, with potential benefits extending beyond Atari to continuous control, multi-task learning, and other domains with complex temporal dependencies.

\bibliography{aaai2026}
\onecolumn   

\appendix             
\setcounter{section}{0} 
\section{Implementation Details}
\label{appendix:reproducibility}

\textbf{Encoder Archictecture.} We adopt the UniZero encoder architecture \cite{unizero}, which builds on the convolutional backbone of LightZero \cite{lightzero} and adds a final linear projection to produce a 768-dimensional (\(D\)) latent state. To improve training stability under partial observability, we incorporate simplicial normalization (SimNorm) \cite{simnorm}, which normalizes each latent segment via a learnable temperature‐controlled mappings.  \\

\noindent\textbf{Transformer Backbone and Prediction Heads.} Our Transformer backbone follows the nanoGPT architecture described in \cite{unizero}, stacking multiple self‐attention and feed‐forward layers to process sequences of timestep inputs. All of our proposed inductive biases are implemented directly within the self‐attention module of each Transformer layer. At each step, the latent state (after SimNorm) and the corresponding action are embedded into a common $D$‐dimensional space via learnable \texttt{nn.Embedding} (or a linear layer for continuous actions) and summed with learnable positional embeddings. The Transformer outputs context‐enriched representations that are sent to two separate two‐layer MLPs with GELU \cite{gelu} activations: the dynamics head predicts the next latent state (of dimension $D$, followed by SimNorm) and the reward distribution (discrete support size), while the decision head predicts policy logits (action‐space size) and value distribution (support size). \\

\noindent\textbf{Training Details. } All reported results are averaged over 5 random seeds, with error bars as described in \autoref{appendix:learned}. Atari environments are provided through the ALE interface (Gymnasium 0.28, sticky actions enabled), ensuring consistency with prior work. All experiments were conducted with a configuration of a single NVIDIA Tesla A100 / V100 GPU, $15-20$ CPU cores, and $60-80$ GB of total RAM. Training an Atari agent for $100{,}000$ environment steps requires approximately $4-5$ hours, with agent evaluations every $10{,}000$ steps (starting after the $20{,}000$th step). We observed stable results across A100 and V100 GPUs. Training configurations can be found in the \texttt{zoo/atari/config} directory, where each attention model has a different configuration file within UniZero. See README file in the codebase for details on how to train an agent.

\noindent\textbf{Compute and Memory Overhead.}  
All proposed priors incur negligible overhead, with at most a 0.002\% increase in MFLOPs per forward pass. \autoref{tab:overhead} shows that parameter counts and FLOPs remain effectively unchanged relative to UniZero, demonstrating that the efficiency gains of adaptive and Gaussian attention come at no meaningful computational cost. 

\begin{table}[h!]
\centering
\scriptsize
\captionsetup{skip=8pt}
\setlength{\tabcolsep}{2pt}
\renewcommand{\arraystretch}{0.8}
\caption{\textbf{Overhead analysis.} Parameter counts (in millions), MFLOPs per Transformer forward pass, and relative increase over the vanilla UniZero baseline.}
\begin{tabular*}{\columnwidth}{@{\extracolsep{\fill}} l r r r r }
\toprule
\textbf{Model} & \textbf{Total Parameters (M)} & \textbf{Transformer Parameters (M)} & \textbf{MFLOPs} & \(\Delta\) MFLOPs (\%) \\
\midrule
Baseline           & 20.77 & 14.18 & 454.611 & —      \\
Adaptive           & 20.77 & 14.18 & 454.615 & \(\bm{+0.001}\)  \\
Gaussian           & 20.77 & 14.18 & 454.619 & \(\bm{+0.002}\)  \\
Gaussian Adaptive  & 20.77 & 14.18 & 454.619 & \(\bm{+0.002}\)  \\
\bottomrule
\end{tabular*}
\label{tab:overhead}
\end{table}

\noindent\textbf{Hyperparameters and Environments.} \autoref{tab:hyperparams} summarizes all architectural and training parameters used in our experiments.  Most values such as latent dimension, Transformer depth, MCTS settings, and optimizer configuration are inherited from UniZero \cite{unizero}, with additional entries for our attention‐bias hyperparameters. All Atari environments are provided through the ALE interface via Gymnasium v0.28, using the standard \texttt{NoFrameskip} variants with sticky actions enabled, matching the settings in the UniZero framework. We select the environments from the Atari 100k benchmark \cite{atari100k}.\\

\begin{longtable}{@{} p{0.45\textwidth} p{0.50\textwidth} @{}}
  \caption{\textbf{Key Hyperparameters}. The values are aligned with those in \cite{unizero} for Atari environments. The section on \textbf{Attention} refers to the newly added parameters.}
  \label{tab:hyperparams}
  \\[-1ex]
  \toprule
  \textbf{Hyperparameter} & \textbf{Value} \\
  \midrule
  \endfirsthead

  \multicolumn{2}{@{}l}{\textbf{(continued)}} \\ 
  \toprule
  \textbf{Hyperparameter} & \textbf{Value} \\
  \midrule
  \endhead

  \multicolumn{2}{@{}l}{\textbf{\textcolor{blue}{Planning}}} \\
  \midrule
  Number of MCTS Simulations (sim) 
    & 50 \\
  Inference Context Length ($H_{\mathrm{infer}}$) 
    & 4 \\
  Temperature 
    & 0.25 \\
  Dirichlet Noise ($\alpha$) 
    & 0.3 \\
  Dirichlet Noise Weight 
    & 0.25 \\
  Coefficient $c_{1}$ 
    & 1.25 \\
  Coefficient $c_{2}$ 
    & 19652 \\
  \midrule

  \multicolumn{2}{@{}l}{\textbf{\textcolor{blue}{Environment and Replay Buffer}}} \\
  \midrule
  Replay Buffer Capacity 
    & 1,000,000 \\
  Sampling Strategy 
    & Uniform \\
  Observation Shape (Atari) 
    & (3, 64, 64) (stack1) \\
  Reward Clipping 
    & True \\
  Number of Frames Stacked 
    & 1 (stack1) \\
  Frame Skip 
    & 4 \\
  Game Segment Length 
    & 400 \\
  Data Augmentation 
    & False \\
  \midrule

  \multicolumn{2}{@{}l}{\textbf{\textcolor{blue}{Architecture}}} \\
  \midrule
  Latent State Dimension ($D$) 
    & 768 \\
  Number of Transformer Heads 
    & 8 \\
  Number of Transformer Layers ($N$) 
    & 2 \\
  Dropout Rate ($p$) 
    & 0.1 \\
  Activation Function 
    & LeakyReLU (encoder); GELU (others) \\
  Reward/Value Bins 
    & 101 \\
  SimNorm Dimension ($V$) 
    & 8 \\
  SimNorm Temperature ($\tau$) 
    & 1 \\
  \midrule

  \multicolumn{2}{@{}l}{\textbf{\textcolor{blue}{Optimization}}} \\
  \midrule
  Training Context Length ($H$) 
    & 10 \\
  Replay Ratio 
    & 0.25 \\
  Buffer Reanalyze Frequency 
    & 1/50 \\
  Batch Size 
    & 64 \\
  Optimizer 
    & AdamW \cite{adamw}\\
  Learning Rate 
    & $1\times 10^{-4}$ \\
  Next Latent State Loss Coefficient 
    & 10 \\
  Reward Loss Coefficient 
    & 1 \\
  Policy Loss Coefficient 
    & 1 \\
  Value Loss Coefficient 
    & 0.5 \\
  Policy Entropy Coefficient 
    & $1\times 10^{-4}$ \\
  Weight Decay 
    & $10^{-4}$ \\
  Max Gradient Norm 
    & 5 \\
  Discount Factor 
    & 0.997 \\
  Soft Target Update Momentum 
    & 0.05 \\
  Hard Target Network Update Frequency 
    & 100 \\
  Temporal Difference (TD) Steps 
    & 5 \\
  Evaluation Frequency 
    & 10k Collector Steps \\
  \midrule

  \multicolumn{2}{@{}l}{\textbf{\textcolor{blue}{Attention}}} \\
  \midrule
  Attention Type
    & \textit{causal}, \textit{gaussian}, \textit{adaptive} or \textit{gaam} \\

 Rotary Positional Embeddings   
    & False \\
  \begin{tabular}[t]{@{}l@{}}
    Initial Gaussian Mean Offset \(\mu^h_0\) \\
    (\texttt{init\_adaptive\_mu})
  \end{tabular}
      & 6.0 (Varied across ablations)\\

  \begin{tabular}[t]{@{}l@{}}
    Initial Gaussian Standard Deviation \(\sigma^h_0\) \\
    (\texttt{init\_adaptive\_sigma})
  \end{tabular}
      & 1.0 (Varied across ablations) \\

   \begin{tabular}[t]{@{}l@{}}
    Max Adaptive Span \\
    (\texttt{max\_adaptive\_span})
  \end{tabular}
      & 20.0 \\

    \begin{tabular}[t]{@{}l@{}}
    Initial Adaptive Span \(L_h^0\) \\
    (\texttt{init\_adaptive\_span})
  \end{tabular}
      & 6.0 (Adaptive), 10.0 (Gaussian Adaptive) \\

    \begin{tabular}[t]{@{}l@{}}
    Adaptive Span Regularization Parameter \\
    (\texttt{adapt\_span\_loss})
  \end{tabular}
      & 0.025 \\

    \begin{tabular}[t]{@{}l@{}}
    Adaptive Span Ramp \(R\) \\
    (\texttt{adapt\_span\_ramp})
  \end{tabular}
      & 3.0\\
  \bottomrule
\end{longtable}
\appendix
\newpage
\setcounter{section}{1}
\section{Learning Curves and Learned Priors}
\label{appendix:learned}

\begin{figure}[h!]
    \centering
    \includegraphics[width=1\linewidth]{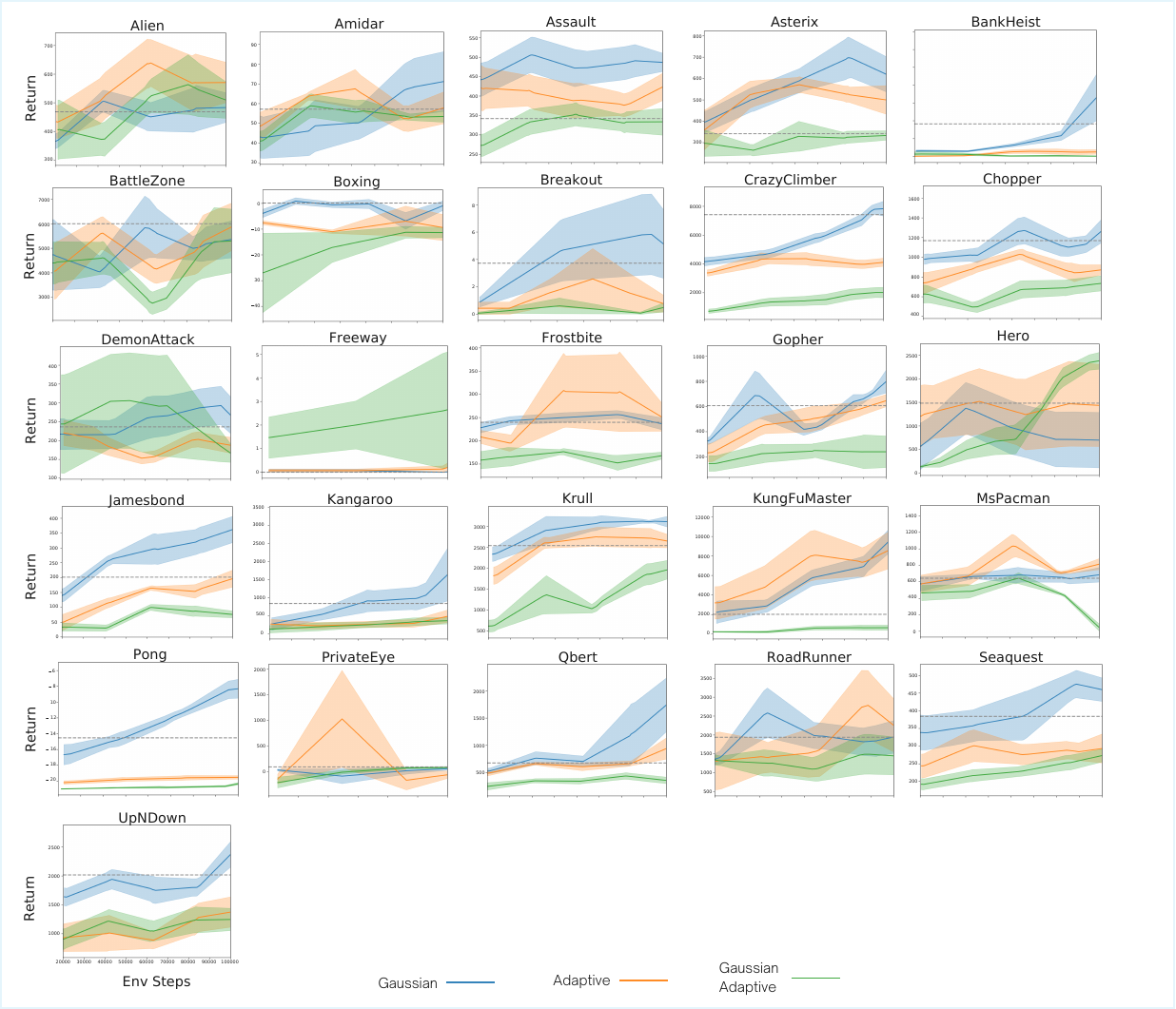}
    \caption{\textbf{Learning Curves for Attention‐Biased UniZero.}  Each panel plots the mean evaluation return (solid line) and standard error (shaded band) over five random seeds for three variants: Gaussian attention (blue), Adaptive attention (orange), and Gaussian Adaptive attention (green). The grey dotted horizontal line in each subplot marks the UniZero baseline’s final return at the 100{,}000th environment step.}
    \label{fig:placeholder}
\end{figure}

\begin{figure}
    \centering
    \includegraphics[width=1\linewidth]{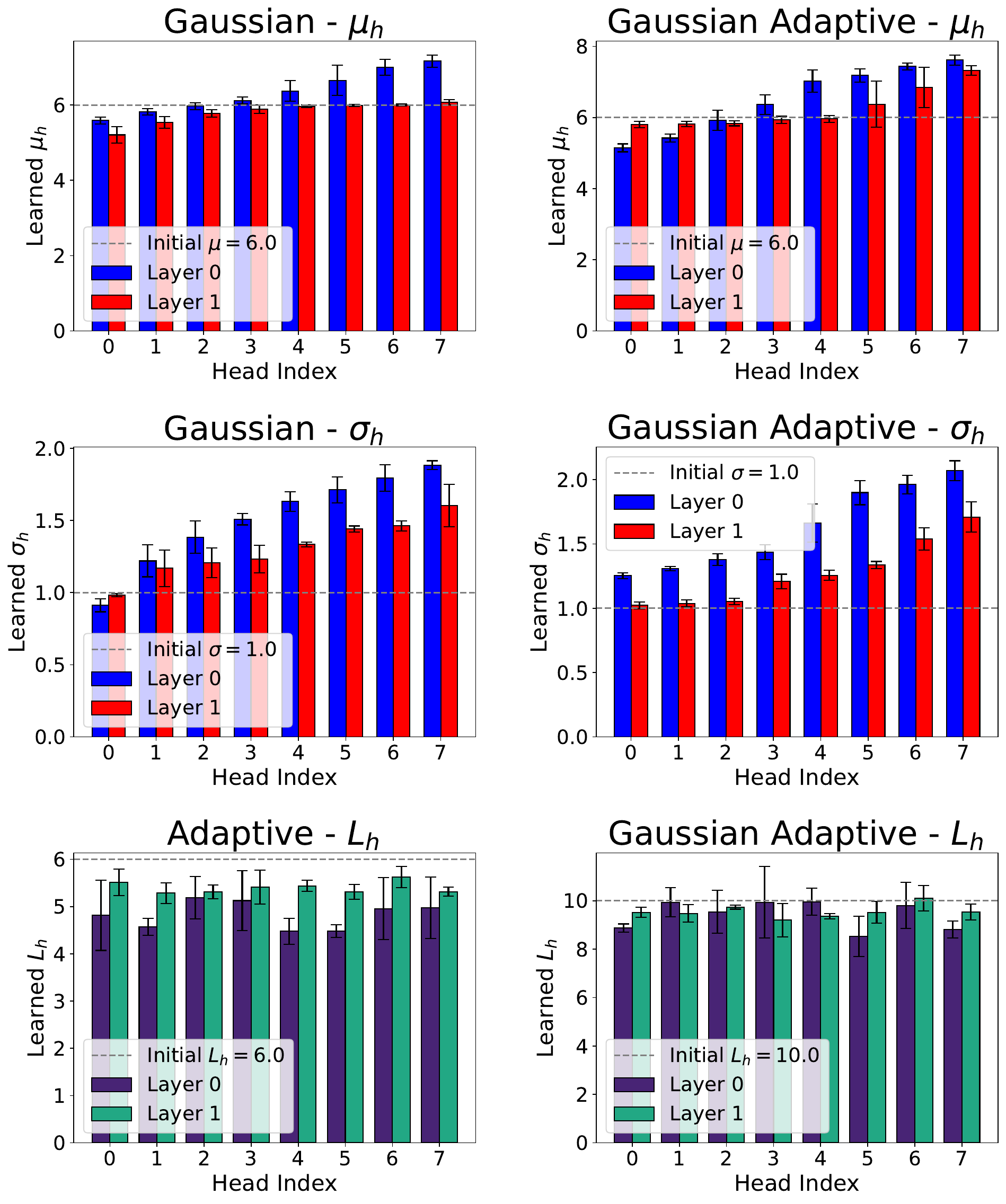}
    \caption{\textbf{Learned adaptive and Gaussian-based attention parameters in Pong.}  The six subplots report the learned values across attention heads and layers, compared against their initialization (dashed lines). 
    Top row: learned Gaussian mean offsets (\(\mu_h\)) for Gaussian (left) and Gaussian Adaptive (right) attention. 
    Middle row: learned Gaussian standard deviations (\(\sigma_h\)) for Gaussian (left) and Gaussian Adaptive (right) attention. 
    Bottom row: learned adaptive memory lengths (\(L_h\)) for Adaptive (left) and Gaussian Adaptive (right) attention. 
    Each bar shows the mean over 5 random seeds, with error bars indicating standard deviations. 
    These plots illustrate how different inductive biases (Gaussian, Adaptive, and Gaussian Adaptive) evolve during training and how learned parameters adapt relative to their initial values.}
    \label{fig:learned}
\end{figure}

In Pong, the learned parameters reveal clear differences between the inductive priors (\autoref{fig:learned}).  

\textbf{Adaptive attention.} The learned memory spans $L_h$ (initialized at 6) drift inconsistently across heads and layers. Some collapse to very short horizons, while others expand far beyond the relevant dependency range. This instability indicates that Adaptive attention struggles to capture Pong’s narrow but stable temporal dependencies.  

\textbf{Gaussian attention.} By contrast, Gaussian attention learns mean offsets $\mu_h$ that remain close to the initialization ($\mu \approx 6$), while widths $\sigma_h$ expand moderately beyond 1.0. This produces smooth, head-specific kernels that emphasize a few recent steps but still leverage informative tails. These stable parameters align well with Pong’s true dependency horizon and explain the stronger performance of this variant.  

\textbf{Gaussian Adaptive attention.} This mechanism combines both priors, but the hard cutoff imposed by $L_h$ (initialized at 10) often truncates the Gaussian kernel. Although the learned $\mu_h$ and $\sigma_h$ resemble those of Gaussian attention, the span clips the tails, removing the soft weighting needed to capture delayed signals. As a result, Gaussian Adaptive inherits the instability of Adaptive rather than the robustness of Gaussian.

\end{document}